\documentclass[letterpaper, 10 pt, journal, twoside]{IEEEtran}

\usepackage{graphics}           
\usepackage{times}              
\usepackage{amsmath}            
\usepackage{amssymb}            
\usepackage{graphicx}
\usepackage{algorithm}
\usepackage[noend]{algpseudocode}
\usepackage{booktabs}
\usepackage{color}
\definecolor{instructioncolor}{rgb}{.5,.5,.5}

\usepackage[font=small]{caption}

\def\secref#1{Sec.~\ref{#1}}
\def\figref#1{Fig.~\ref{#1}}
\def\tabref#1{Tab.~\ref{#1}}
\def\eqref#1{Eq.~(\ref{#1})}

\makeatletter
\usepackage{xspace}
\DeclareRobustCommand\onedot{\futurelet\@let@token\@onedot}
\def\@onedot{\ifx\@let@token.\else.\null\fi\xspace}
\def\eg{e.g\onedot} 
\def\ie{i.e\onedot}

\def\wrt{w.r.t\onedot} 
\def\etal{{et al}\onedot}
\makeatother

\def\etalcite#1{\etal~\cite{#1}}

\usepackage{array}
\newcolumntype{L}[1]{>{\raggedright\let\newline\\\arraybackslash\hspace{0pt}}m{#1}}
\newcolumntype{C}[1]{>{\centering\let\newline\\\arraybackslash\hspace{0pt}}m{#1}}
\newcolumntype{R}[1]{>{\raggedleft\let\newline\\\arraybackslash\hspace{0pt}}m{#1}}

\newcommand{\RR}{\mathbb{R}}

\newcommand{\norm}[1]{\lVert#1\lVert}

\usepackage{multirow}
\usepackage{hyperref}
\usepackage[dvipsnames]{xcolor}

\title{IR-MCL: Implicit Representation-Based\\ Online Global Localization}

\author{Haofei Kuang \quad Xieyuanli Chen \quad Tiziano Guadagnino \quad Nicky Zimmerman \quad Jens Behley \quad Cyrill Stachniss
	\thanks{Manuscript received: Sep 7, 2022; Revised: Dec 2, 2022; Accepted: Jan 10, 2023. This paper was recommended for publication by Editor Sven Behnke upon evaluation of the Associate Editor and Reviewers' comments.}%
	\thanks{This work has partially been funded by the European Union’s Horizon 2020 research and innovation programme under grant agreement No~101017008~(Harmony). All authors are with the University of Bonn, Germany. Cyrill Stachniss is additionally with the Department of Engineering Science at the University of Oxford, UK, and with the Lamarr Institute for Machine Learning and Artificial Intelligence, Germany.}%
	\thanks{Digital Object Identifier (DOI): see top of this page.}
}

\begin{document}
\maketitle

\markboth{IEEE Robotics and Automation Letters. Preprint Version. Accepted January, 2023}
{Kuang \MakeLowercase{\textit{et al.}}: IR-MCL: Implicit Representation-Based Online Global Localization}

\begin{abstract}
%
Determining the state of a mobile robot is an essential building block of robot navigation systems.
In this paper, we address the problem of estimating the robot's pose in an indoor environment using 2D LiDAR data and investigate how modern environment models can improve gold standard Monte-Carlo localization~(MCL) systems. 
We propose a neural occupancy field to implicitly represent the scene using a neural network.
With the pretrained network, we can synthesize 2D LiDAR scans for an arbitrary robot pose through volume rendering. 
Based on the implicit representation, we can obtain the similarity between a synthesized and actual scan as an observation model and integrate it into an MCL system to perform accurate localization.
We evaluate our approach on self-recorded datasets and three publicly available ones.
We show that we can accurately and efficiently localize a robot using our approach surpassing the localization performance of state-of-the-art methods.
The experiments suggest that the presented implicit representation is able to predict more accurate 2D LiDAR scans leading to an improved observation model for our particle filter-based localization.
The code of our approach will be available at: \url{https://github.com/PRBonn/ir-mcl}.
\end{abstract}

\section{Introduction}
\label{sec:intro}
\IEEEPARstart{L}{ocalizing} a robot on a known map is a key capability often needed by mobile robots deployed in indoor environments.
For such indoor localization, we often need a map representation of the scene to establish an observation model to correct the pose estimate of a probabilistic localization algorithm, such as Monte-Carlo localization~(MCL)~\cite{dellaert1999icra}. The map representation quality and the observation model's design are critical for localization accuracy.

Recently, learning-based methods are widely used in the computer vision domain for representing the surrounding~\cite{mildenhall2020eccv, park2019cvpr, peng2020eecv}. 
Among these works, Mildenhall~\etalcite{mildenhall2020eccv} propose the seminal work of neural radiance fields~(NeRF), which learns an implicit function to encode the environment that can be used to generate novel views at new poses using volumetric rendering.
The generated views show a high fidelity including direction-dependent illumination effects and attracted increasing interest in the computer vision community. 
Moreover, the implicit representation encoded by a neural network has appealing properties also relevant for robotics applications: 
They offer a compact representation that only needs to store the parameters of the trained neural network, and they can generalize well to locations not seen during the training.
Recently, multiple works~\cite{deng2022cvpr, li2021iccv, oechsle2021iccv, sucar2021iccv, zhu2022cvpr} have been proposed to leverage depth information to impose stronger geometric constraints. 
In this work, we are interested in global localization using 2D LiDAR sensors commonly employed in indoor robotics.
In indoor environments, occupancy grid maps~\cite{thrun2005probrobbook} are widely used to explicitly represent the environment.
However, the discrete nature of occupancy maps can cause loss of scene details, which potentially leads to an inaccurate observation model of probabilistic localizations algorithms~\cite{dellaert1999icra} and consequently inaccurate localization results.


\begin{figure}[t]
\centering
\includegraphics[width=1.0\linewidth]{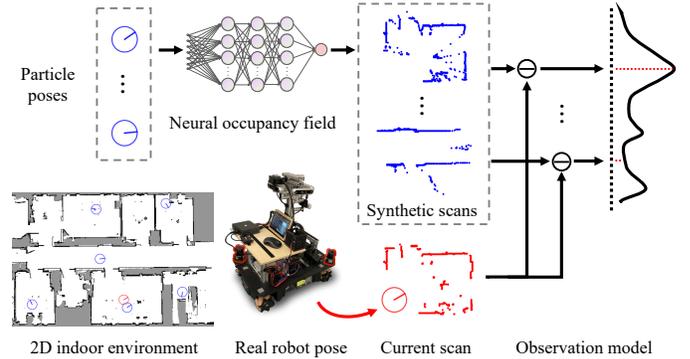}
\caption{Given a set of particles and a real scan from 2D LiDAR, we establish an observation model with a pretrained neural representation for accurate robot global localization.}
\label{fig:motivation}
\vspace{-0.6cm}
\end{figure}

The main contribution of this paper is the use of an implicit NeRF-based representation of the environment for MCL together with an observation model exploiting this implicit representation. 
It tackles the limitation of the discrete occupancy grid map and improves the localization accuracy.
Our proposed method represents the 2D world using an implicit function through a neural occupancy field, named NOF.
It exploits a multi-layer perceptron~(MLP) to encode the 2D world. Given a location, the MLP outputs the corresponding occupancy probability. 
Based on that, our method then uses a ray casting-based rendering algorithm to synthesize a range scan for an input sensor pose, see~\figref{fig:motivation} for an illustration.
We train the NOF by comparing the rendered synthetic scan to the real sensor measurements. 
We use the NOF to build a novel observation model for MCL~\cite{dellaert1999icra}. For each particle in MCL, we use our NOF to render a synthetic view and compare it to the current observation to update the particle weight. 
We call our global localization system implicit representation-based MCL~(IR-MCL).

In summary, we make the following three key claims: 
(i)~we are able to build an effective observation model based on the proposed implicit representation of the environment for 2D LiDAR-based (global) localization;
(ii)~we achieve state-of-the-art localization performance compared to approaches using occupancy grid maps;
(iii)~our approach converges fast to globally localize a robot and operates online.
We support these claims by our experimental evaluation on multiple datasets.

\section{Related Work}
\label{sec:related}

\begin{figure*}[t]
\centering
\vspace{0.2cm}
\includegraphics[width=1.0\linewidth]{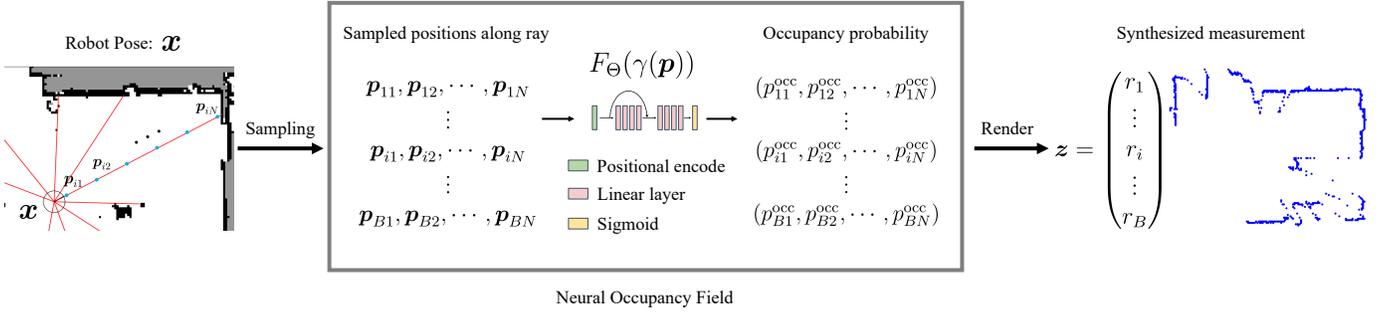}
\caption{Overview of our approach for rendering a synthesized measurement of LiDAR from our implicit scene representation model: Neural Occupancy Field (NOF). 
We uniformly sample multiple positions along each LiDAR beam, our NOF is a neural network that takes 2D position $ \boldsymbol{p} = (x, y) $ as input and outputs an occupancy probability of it, we can synthesize range values with all predictions along LiDAR beams through volume rendering.}
\label{fig:overview}
\vspace{-0.6cm}
\end{figure*}

For global localization and pose tracking, Dellaert~\etalcite{dellaert1999icra} propose using particle filters to realize Monte-Carlo localization. 
MCL is still the gold standard for robot localization and often uses LiDARs~\cite{dellaert1999icra,fox2001neurips,stachniss2005aaai,yan2019ecmr}, cameras~\cite{dellaert1999icra,bennewitz2006euros}, or WiFi~\cite{ito2014icra}. Fox~\etalcite{fox2001neurips} propose an adaptive sampling strategy for MCL to significantly improve its efficiency.
Yilmaz~\etalcite{yilmaz2019ras} propose self-adaptive MCL which is improved to make the algorithm suitable for autonomous  vehicles.
Such MCL methods often use a 2D LiDAR sensor and an occupancy grid map to estimate the robot's pose and are quite robust.

Recent learning-based localization algorithms also achieve high precision global localization.
Lu \etalcite{lu2019cvpr} propose L$^{3}$Net, which optimize a deep neural network to optimize the robot's pose using a 3D LiDAR scan with a pre-built 3D point cloud map.
L$^{3} $Net achieves centimeter-level accuracy in an urban environment but needs as prior an estimate of the robot's pose.
This approach is often referred to as pose tracking. 
Chen \etalcite{chen2020iros} exploit a convolutional neural network to predict the overlap between a real LiDAR scan and a virtual scan from the pre-build map, which is used as an observation model in an MCL framework.
Zimmerman \etalcite{zimmerman2022iros} combine 2D LiDAR-based localization using occupancy grid maps in the MCL framework and text spotting using an additional camera to enhance the robustness of the indoor localization. 

Our proposed IR-MCL approach exploits an implicit representation of the environment to model the scene and define the observation model for localization.
The classic map representation used in robot localization~\cite{chen2021icra, dellaert1999icra, fox1999aaai, stachniss2003ijcai} is the discrete occupancy grid map, which is limited by the resolution of grid cells that loses the detailed geometric information of the scene.
To cope with this challenge, continuous Gaussian process grid maps~\cite{o2012ijrr, yuan2018icarcv}, Hilbert maps~\cite{ramos2016ijrr}, and a feature-based implicit representations~\cite{zhao2020ral} have been proposed to represent the 2D world. 
Often these functions may not generalize well for describing the world precisely and therefore can limit the global localization capabilities.

In recent years, deep learning-based methods for representing the environment are widely used in the computer vision.
Mildenhall~\etalcite{mildenhall2020eccv} propose a method to learn an implicit function to represent the scene by modeling a neural radiance field. It can predict realistic scene-aware views for arbitrary input poses to support many applications such as virtual/augmented reality or robot navigation using cameras~\cite{adamkiewicz2022ral}.

In contrast to the vanilla NeRF that predicts the volume density, 
Xu~\etalcite{xu2021neurips} propose a generative occupancy field to represent surfaces by predicting the occupancy probability of the space.
Similarly, Oechsle~\etalcite{oechsle2021iccv} propose a volume ray-tracing algorithm for the occupancy field to render surface-aware images.
Based on the occupancy fields, several works exploit the depth information as a strong geometry constraint from RGB-D sensors~\cite{sucar2021iccv,zhu2022cvpr}, or depth estimation algorithms~\cite{deng2022cvpr,li2021iccv,rematas2022cvpr} together with color information to train NeRF for rendering of novel depth images or even point clouds.
For example, UrbanNeRF by Rematas~\etalcite{rematas2022cvpr} train a large-scale NeRF model with high-precision 3D LiDAR to perform more realistic 3D reconstruction at city-scale.
These works suggest that geometric information supports the learning of NeRFs to obtain appealing performance. 
Beyond the NeRF, iSDF~\cite{ortiz2022rss} and CNM~\cite{yan2021iccv} learn a signed distance function (SDF) through a neural network as the map representation to trade-off between accuracy and efficiency.
LASER~\cite{min2022cvpr} exploits the latent space for robotic visual localization.

Recently, neural implicit representations have also been used to support robot mapping and localization.
For example, iNeRF by Lin~\etalcite{lin2021iros} estimates the camera pose by inverting the training process of NeRF to optimize the camera pose with a trained NeRF.
There are also works that exploit an implicit scene representation in localization and mapping for mobile robots.
Moreau~\etalcite{moreau2021corl} propose to use NeRF to synthesize observations and enhance the mapping and localization results under limited amount of real data.
They later also propose ImPosing~\cite{moreau2023wacv}, which uses the implicit representation to achieve real-time loop closing at city-scale.
Adamkiewicz~\etalcite{adamkiewicz2022ral} build a vision-only navigation system based on a pre-trained NeRF to forecast the measurement of the future robot state for optimizing the trajectory. 
Concurrent to our work, Loc-NeRF~\cite{maggio2022arxiv} was released, which exploits the implicit map representation for visual localization.
To the best of our knowledge, our proposed IR-MCL system is the first work that uses an implicit neural representation as an observation model for LiDAR-based global localization in indoor environments.


\section{Approach}
\label{sec:main}

To realize IR-MCL, we study the problem of generating 2D LiDAR scans at arbitrary sensor positions in a scene through a neural implicit representation for robot global localization.
To this end, we propose a neural network to predict the occupancy probability for a given location to represent a 2D environment as detailed in~\secref{subsec:nof}.
Based on such estimated occupancy probabilities of samples along LiDAR rays, we render a synthetic LiDAR scan for a given pose of the robot as presented in~\secref{subsec:rendering}.
Compared with the real measurements from 2D LiDAR during training, we optimize the weights of the network as described \mbox{in~\secref{subsec:training}}. 
After that, we use the trained network to build a novel observation model and integrate it into the MCL framework to achieve efficient global localization as presented in~\secref{subsec:mcl}.
\figref{fig:overview} shows an overview of our method.

\subsection{An Implicit Representation: Neural Occupancy Field}
\label{subsec:nof}
We propose a neural network to predict the occupancy probability of an input 2D location $\boldsymbol{p} \in \RR^2$ as the implicit scene representation, named neural occupancy field or NOF in short.
Our approach uses a function $ F_\Theta $ to implicitly represent a continuous 2D world. 
More specifically, This function takes a 2D location $\boldsymbol{p} = (x, y)^{\top}$ as input and outputs the corresponding occupancy probability $ p^{\text{occ}}$ as:
\begin{align}
p^{\text{occ}} &= F_\Theta(\gamma(\boldsymbol{p})).
\label{eq:nof}
\end{align}

We represent $F_\Theta$ using an MLP inspired by NeRF~\cite{mildenhall2020eccv}, where $\Theta$ represents the weights of the neural network. 
In line with NeRF, we also use positional encoding to project a 2D location to a high-dimensional space to encourage our model to encode higher frequency information of the world.
We use $\gamma(\boldsymbol{p})$ with the positional encoding:
\begin{align}
\gamma(\boldsymbol{p})~=~&[\boldsymbol{p}, \sin(2^{0}\boldsymbol{p}), \cos(2^{0}\boldsymbol{p}), \dots, \nonumber \\
&\sin(2^{L-1}\boldsymbol{p}), \cos(2^{L-1}\boldsymbol{p})],
\label{eq:sampling}
\end{align}
where we use $L = 10$ in our implementation.

The network is trained such that it can map from arbitrary input 2D coordinates to the corresponding occupancy probability.
To accomplish this, our MLP consisted of 8 fully-connected layers, each followed by batch normalization~\cite{ioffe2015icml} and a ReLu activation. 
Additionally, we adopt and include residual connections~\cite{he2016cvpr} to improve the accuracy of the predictions.
We apply an additional fully-connected layer followed by a sigmoid activation on the output $ D $-dimensional feature vector generated by the MLP to obtain the occupancy probabilities~$ p^{\text{occ}} \in [0, 1]$.

Our network predicts an occupancy probability $ p^{occ}~\in~[0, 1] $, which can be used for representing the 2D scene.
That is different from existing neural representations, such as NeRF~\cite{mildenhall2020eccv}, which represents the scene geometry from the predicted volume density. Our proposed network requires no threshold adjustment to get the occupancy state (free or occupied). Thus, it generalizes well to different scenes.

\subsection{Novel View Rendering with NOF}
\label{subsec:rendering}
Based on the proposed NOF representation, we can render a novel LiDAR scan for an arbitrary 2D pose in the environment through the ray casting algorithm.

More specific, given a current 2D pose $ \boldsymbol{x} = (x ,y, \theta)^{\top} $ of a robot, we determine the origin $\boldsymbol{o} = (x, y)^{\top}$ and the normalized direction vector $\boldsymbol{d} = (d_1, d_2)^{\top}$ of each LiDAR beam. 
The direction vector of a ray $ \boldsymbol{d} $ is calculated from the robot orientation $ \theta $ and the parameters of the 2D LiDAR sensor. 
We uniformly sample $N$ points $\boldsymbol{p}_i = \boldsymbol{o} + m_i \boldsymbol{d}$ along the ray, where $m_i$ is the distance from the origin $ \boldsymbol{o} $ to the sampled point $\boldsymbol{p}_i$ limited by the valid measurement range of the 2D LiDAR sensor, \ie, $m_i \in [m_{\min}, m_{\max}]$.
Similar to prior work~\cite{rematas2022cvpr}, we model the termination weights $\alpha_{i}$ at the endpoint $\boldsymbol{p}_i$ along the ray as:
\begin{align}
\alpha_{i} &= p_{i}^{\text{occ}} \prod_{j = 1}^{i-1} (1 - p_{j}^{\text{occ}}),
\label{eq:likelihood_weights}
\end{align}
where we assume that all occupancy probabilities~$p_i^{\text{occ}}$ are independent.
With this, we can compute a range $r \in \RR$ according to the termination weights of the samples $\boldsymbol{p}_i$ and their distances $m_{i}$ along the ray by:
\begin{align}
r &= \sum_{i = 1}^{N} \alpha_{i} m_{i}.
\label{eq:rendering_range}
\end{align}

Repeating this procedure for each LiDAR beam, we can render a synthetic observation at any query location $\boldsymbol{x}$ based on our NOF, and use this scan in comparison to the real scan for the MCL observation model.

\subsection{Training the NOF}
\label{subsec:training}
Based on the above-introduced rendering algorithm, we can train the neural network $F_\Theta$ using recorded 2D LiDAR scans and the corresponding pose as done when building the map for traditional MCL. 
Each scan $ \boldsymbol{z}_t$ at time $t$ is recorded from a pose $ \boldsymbol{x}_t = (x, y, \theta)^{\top}_t$.
According to the parameters of the LiDAR sensor, each scan is comprised of $ B $ beams and each beam corresponds to a real range value $\hat{r}_i \in \RR$ of the $i^\text{th}$ beam.
We use two loss functions for optimizing the weights~$\Theta$ of our MLP network, a geometric loss, and an occupancy regularization.

\subsubsection{Geometric Loss}
We compute the geometric loss between the rendered range value~$r_i$ and the recorded range value~$\hat{r}_i$ of the $i^\text{th}$ beam using the $L_{1}$ loss:
\begin{align}
\mathcal{L}_{geo} &= \frac{1}{B}\sum_{i = 1}^{B}|{r_{i} - \hat{r}_i}|.
\label{eq:loss_geo}
\end{align}

We opt for using the $L_{1}$ instead of the $L_2$ loss to reduce the influence of the measurement noise of the employed 2D LiDAR sensor.

\subsubsection{Occupancy Regularization}
The predicted value in NOF is regarded as the occupancy value at the input location. Therefore, $ p^{\text{occ}} $ is expected to be equal to $ 1 $ for the occupied space and $ 0 $ for the free space. That means, ideally, the entropy of the prediction should be $ 0 $. Following similar work~\cite{xu2021neurips}, we add a negative log-likelihood loss as regularization to reduce the entropy of predicted occupancy probabilities:
\begin{align}
\mathcal{L}_{reg} &= \frac{1}{N \, B} \sum_{i=1}^{N \, B}\log(F_\Theta(\boldsymbol{p}_i)) + \log(1 - F_\Theta(\boldsymbol{p}_i)),
\label{eq:loss_reg}
\end{align}
where $ N $ is the number of sampled points along each beam.

The final loss function is then given by:
\begin{equation}
\mathcal{L} = \mathcal{L}_{geo} + \lambda\mathcal{L}_{reg},
\label{eq:loss}
\end{equation}
where $ \lambda $ is a hyperparameter to balance the influence of the occupancy regularization.

We train our NOF network using the Adam optimizer~\cite{kingma2015iclr} with a batch size of $ 1024 $ for all datasets in all experiments.
During rendering, we sample $ 256 $ points for each LiDAR beam, \eg~$ N = 256 $, and train the network for $ 32 $ epochs.
The initial learning rate is $ 10^{-4} $ and decayed by $ 0.5 $ at epoch $ 4 $ and epoch $ 8 $, and weight decay is $ 0.001 $. 
The balancing coefficient of occupancy regularization is set to $ \lambda = 10^{-5} $.

\subsection{Implicit Representation MCL (IR-MCL)}
\label{subsec:mcl}
Based on the rendered observations by our NOF network, we propose a novel observation model for MCL to achieve global localization. 
The global localization is formulated as a posterior probability estimation problem~\cite{thrun2005probrobbook}, where the objective is to estimate the belief $ bel(\boldsymbol{x}_t) $ at the robot's pose $ \boldsymbol{x}_t = (x, y, \theta)^{\top}_t$ at time $ t $. 
The update of the belief $ bel(\boldsymbol{x}_t) $ uses a recursive Bayes filter and is formulated as:
\begin{equation}
bel(\boldsymbol{x}_t) = \eta\, p(\boldsymbol{z}_t \mid \boldsymbol{x}_t, \mathcal{M}) \, \overline{bel}(\boldsymbol{x}_t),
\label{eq:baysian_filter}
\end{equation}
where $ \overline{bel}(\boldsymbol{x}_t) $ is the predicted belief of the robot pose according to the motion controls and the last pose $ \boldsymbol{x}_{t-1} $, which is also called motion model of the robot. 
The \mbox{$ p(\boldsymbol{z}_t \mid \boldsymbol{x}_t, \mathcal{M}) $} is the likelihood of the sensor measurement $ \boldsymbol{z}_t $ while the robot state is $ \boldsymbol{x}_t $ in the map $ \mathcal{M} $.
It is also regarded as the observation model for correcting the estimate of motion model.
The $ \eta $ is a normalization factor.

\begin{figure}[t]
\centering
\vspace{0.05cm}
\includegraphics[width=1.0\linewidth]{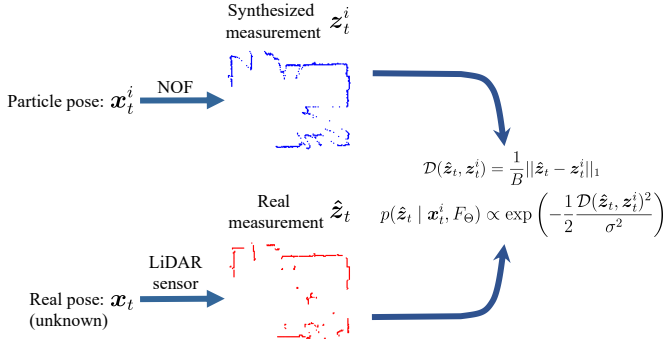}
\caption{The implicit representation-based observation model. We render a 2D LiDAR scan for each particle, then update the weights of the particle by comparing the synthesized measurement with the real measurement from the LiDAR sensor.}
\label{fig:observation_model}
\vspace{-0.6cm}
\end{figure}

MCL exploits a particle filter to approximate the update of posterior $ bel(\boldsymbol{x}_t) $ by a set of random samples drawn from the posterior. These random samples, so-called particles, denoted as $ \mathcal{X}_t = \{(\boldsymbol{x}_{t}^{1}, w_{t}^{1}), (\boldsymbol{x}_{t}^{2}, w_{t}^{2}), \dots, (\boldsymbol{x}_{t}^{M}, w_{t}^{M})\} $, where $ w^{i} $ is the weight of the pose $ \boldsymbol{x}^{i} $, and $ M $ is number of particles.
After updating, the particles are re-sampled according to the particles' importance weights.
Repeating this process, the particles eventually converge to a small region around the real pose.

\begin{table*}[t]
	\centering
	\renewcommand\arraystretch{0.95}
	\scalebox{1.00}{
		\begin{tabular}{c|c|c|cccc|cccc}
			\toprule
			\multirow{2}{*}{Sequence}               & \multirow{2}{*}{Method} 		& \multirow{2}{*}{Memory}        & Location & $ <5 $cm & $ <10 $cm & $ <20 $cm & Yaw  & $ <0.5^\circ $ & $ <1^\circ $ & $ <2^\circ $ \\
			& 		& 		& RMSE (cm) $\downarrow$ & Pct. $\uparrow$ & Pct. $\uparrow$ & Pct. $\uparrow$  & RMSE (degree) $\downarrow$ & Pct. $\uparrow$ & Pct. $\uparrow$ & Pct. $\uparrow$ \\
			\midrule
			\multirow{5}{*}{Seq 1} & AMCL 		&\multirow{3}{*}{4\,MB}        & 11.57             & 24.44\%                    & 58.89\%                    & 92.22\%                     & 1.80         & 21.11\%                        & 44.44\%                     & 81.11\%                      \\
			& NMCL 		&      & -              & 17.36\%                    & 32.98\%                    & 81.71\%                     & 7.14         & 22.16\%                        & 39.12\%                     & 67.16\%                      \\
			& SRRG-Loc 		&       & 6.36              			 & 49.11\%                    			   & 92.10\%                    			 & 99.85\%                       			  & 1.08          				& \textbf{47.25\%}                        			 & \textbf{75.45\%}                     				& 94.00\%                      \\
			\cline{3-3}
			& HMCL 		& 0.01\,MB     & 13.44              & 18.46\%                    & 32.98\%                    & 81.71\%                     & 3.33         & 19.57\%                        & 38.55\%                     & 67.50\%                      \\
			& IR-MCL 		 & 1.96\,MB 		& \textbf{5.13}              & \textbf{63.15\%}                   & \textbf{97.27\%}                    & \textbf{100.00\%}                     & \textbf{1.05}          & 47.12\%                       & 74.59\%                     & \textbf{94.24\%}                      \\
			\midrule
			\multirow{5}{*}{Seq 2} & AMCL 		&\multirow{3}{*}{4\,MB}          & 10.65              & 17.11\%                    & 52.63\%                    & 98.68\%                     & 1.09         & 28.95\%                        & 55.26\%                     & 94.74\%                     \\
			& NMCL 		&    & 23.52              & 19.78\%                    & 40.19\%                    & 74.96\%                     & 4.51         & 19.78\%                        & 40.96\%                     & 65.38\%                      \\
			& SRRG-Loc 		&        & 8.83            			  & 25.79\%                    			   & 69.09\%                    			& \textbf{100.00\%}                       			 & 1.43         			  & 27.82\%                        				& 53.54\%                     			   & 82.28\%                      \\
			\cline{3-3}
			& HMCL 		& 0.01\,MB   & -              & 0.00\%                    & 0.00\%                    & 0.00\%                     & -         & 0.00\%                        & 0.00\%                     & 0.00\%                      \\
			& IR-MCL  	& 1.96\,MB 		  & \textbf{5.53}              & \textbf{62.20\%}                   & \textbf{92.85\%}                   & \textbf{100.00\%}                     & \textbf{0.81}         & \textbf{48.88\%}                        & \textbf{82.15\%}                     & \textbf{98.16\%}                      \\
			\midrule
			\multirow{5}{*}{Seq 3} & AMCL 		&\multirow{3}{*}{4\,MB}          & -              & 30.77\%                    & 71.79\%                    & 84.62\%                     & -         & 15.38\%                        & 23.08\%                     & 61.54\%                      \\
			& NMCL 		&    & -              & 0.00\%                    & 0.00\%                    & 0.00\%                     & -         & 0.64\%                        & 0.96\%                     & 1.61\%                      \\
			& SRRG-Loc 		&        & 50.36                    			& 20.91\%                      			   & 39.38\%                    			 & 77.98\%                        			  & 2.86               			 & 17.80\%                        			   & 22.25\%                     			  & 43.38\%                      \\
			\cline{3-3}
			& HMCL 		& 0.01\,MB   & -              & 0.00\%                    & 0.00\%                    & 0.00\%                     & -         & 0.00\%                        & 0.00\%                     & 0.00\%                      \\
			& IR-MCL 		 & 1.96\,MB 		 & \textbf{4.59}              & \textbf{80.42\%}                    & \textbf{98.33\%}                    & \textbf{100.00\%}                     & \textbf{0.65}         & \textbf{60.18\%}                        & \textbf{85.54\%}                     & \textbf{100.00\% }                     \\
			\midrule
			\multirow{5}{*}{Seq 4} & AMCL 		&\multirow{3}{*}{4\,MB}          & -             & 20.55\%                    & 54.79\%                    & 83.56\%                     & 9.57        & 24.66\%                        & 42.47\%                     & 68.49\%                      \\
			& NMCL 		&    & -              & 0.00\%                    & 0.00\%                    & 0.00\%                     & -         & 0.35\%                        & 0.70\%                     & 2.29\%                      \\
			& SRRG-Loc 		&        & \textbf{11.02}              			  & 16.17\%                    			   & 51.73\%                    			 & \textbf{96.04\%}                    & \textbf{1.15}         		       & \textbf{45.79\%}                          & 68.40\%                    				 & \textbf{89.44\%}                      \\
			\cline{3-3}
			& HMCL 		& 0.01\,MB   & 22.70              & 0.50\%                    & 6.77\%                    & 61.1\%                     & 4.42         & 13.28\%                        & 27.39\%                     & 52.23\%                      \\
			& IR-MCL 		 & 1.96\,MB         & 11.54            & \textbf{38.78\%}                & \textbf{67.82\%}                      & 92.74\%                     				& 1.56         & 37.94\%                        			   & \textbf{68.73\%}                        & 84.74\%                      \\
			\midrule
			\multirow{5}{*}{Seq 5} & AMCL 		&\multirow{3}{*}{4\,MB}          & -              & 15.87\%                    & 58.73\%                    & 92.06\%                     & -         & 20.63\%                        & 53.97\%                     & 85.71\%                      \\
			& NMCL 		&    & 47.15              & 7.78\%                    & 45.91\%                     & 84.83\%                      & 3.82         & 22.95\%                        & 39.12\%                     & 52.10\%                      \\
			& SRRG-Loc 		&        & -                             & 34.97\%                                 & 69.85\%                    			  & 94.49\%                        			   & 1.87         			   & \textbf{38.24\%}                        			 & \textbf{73.30\%}                    & \textbf{90.78\%}                      \\
			\cline{3-3}
			& HMCL 		& 0.01\,MB   & 23.97              & 1.46\%                    & 9.99\%                    & 57.62\%                     & 5.76         & 9.30\%                        & 24.46\%                     & 49.87\%                      \\
			& IR-MCL 		 & 1.96\,MB         & \textbf{6.33}                   &\textbf{48.15\%}                    & \textbf{90.27\%}                   & \textbf{100.00\%}                    & \textbf{1.47}         & 37.98\%                        & 60.03\%                     			 & 81.22\% \\          
			\bottomrule           
		\end{tabular}
	}
	\caption{Quantitative results on IPBLab dataset. We compare with various MCL-based global localization algorithms. We report the absolute pose error metrics in location and direction, the `-'  means global localization failed in the sequence. We only report the accuracy for success cases. Alongside, we also report the ratio of frames with location and yaw angle errors less than the given threshold.}
	\label{tab:ipblab_loc}
	\vspace{-0.4cm}
\end{table*}

In this work, we exploit our NOF model to implicitly represent the environment $ \mathcal{M} $ to generate scans used in our observation model for MCL to achieve global localization. 
More specific, our IR-MCL treats each particle as a hypothesized robot pose at time $ t $, \eg $ \boldsymbol{x}^{i}_{t} = (x^i, y^i, \theta^i)^{\top}_{t} $. We render an observation $ \boldsymbol{z}^{i}_{t} $ at each particle location as described in \secref{subsec:rendering}, and compare it with the real measurement $ \boldsymbol{\hat{z}}_t $ obtained by the 2D LiDAR sensor, which is shown in \figref{fig:observation_model}.
Following~\cite{chen2021icra}, we approximated the likelihood \mbox{$ p(\boldsymbol{\hat{z}}_t \mid \boldsymbol{x}_t^{i}, F_{\Theta}) $} of the $ i $-th particles through a Gaussian distribution:
\begin{align}
p(\boldsymbol{\hat{z}}_t \mid \boldsymbol{x}_t^{i}, F_{\Theta}) &\propto \exp \left(-\frac{1}{2} \frac{\mathcal{D}(\boldsymbol{\hat{z}}_t, \boldsymbol{z}^{i}_{t})^{2}}{\sigma^{2}}
\right),
\label{eq:sensor_model}
\end{align}
where $ \mathcal{D}(\boldsymbol{\hat{z}}_t, \boldsymbol{z}^{i}_{t}) $ is the difference between the measurement $ \boldsymbol{\hat{z}}_t $ and $ \boldsymbol{z}^{i}_{t} $. We use the $ L_{1} $ distance to calculate $ \mathcal{D} $, \ie, $ \mathcal{D}(\boldsymbol{\hat{z}}_t, \boldsymbol{z}^{i}_{t}) = \frac{1}{B}\norm{\boldsymbol{\hat{z}}_t - \boldsymbol{z}^{i}_{t}}_{1} $. It is robust to the noise of the measurements and easy to use while also maintaining high efficiency.
By comparing the current real-sensor measurement with the synthetic observations rendered at all particle locations, we update the likelihood $ p(\boldsymbol{\hat{z}}_t \mid \boldsymbol{x}_t, F_\Theta) $.

To accelerate the runtime of our IR-MCL while using a large number of particles, \eg, $ M = 100,000$, 
we build a predefined 2D grid to store the predicted probabilities for accelerating the rendering similar to~\cite{hedman2021iccv}, and call it neural occupancy grids (NOG). The NOG will cover the whole space of the current scene. 
During localization, we use the nearest neighbor cell of each sampling point along the ray in the NOG as the occupancy probability at this point. 
By exploiting NOG, our IR-MCL system achieves real-time performance even with a large number of particles.

\section{Experimental Evaluation}
\label{sec:exp}

The main focus of this work is an implicit representation-based Monte-Carlo localization system for the global localization of a robot.
We present our experiments to show the capabilities of our method and support our key claims, which are:
(i) we are able to build an accurate observation model based on the proposed NOF for 2D LiDAR-based global localization,
(ii) we achieve state-of-the-art localization performance compared to approaches using occupancy grid maps,
(iii) our approach converges fast to globally localize a robot and operates online.

\subsection{Experimental Setup}
\label{subsec:exp_setup}

\textbf{Datasets.} We evaluate our method and compare it with the state-of-the-art methods in multiple datasets including, 
three typical publicly available datasets: Freiburg Building 079 (shortly Fr079) dataset, Intel Lab dataset, MIT CSAIL Lab (shortly MIT), 
and a self-recorded dataset, called IPB\-Lab dataset. 
The three publicly datasets only contain one sequence of an indoor scenario, therefore, we split each sequence into three subsets for training, validation, and testing.
The Fr079 contains $ 3448 $ frames for training, $ 384 $ and $ 959 $ frames for validation and testing respectively.
The Intel Lab dataset contains $ 655 $ frames for training, $ 73 $ and $ 182 $ frames for validation and testing respectively.
The MIT dataset contains $ 291 $ frames for training, $ 33 $ and $ 82 $ frames for validation and testing respectively.
The IPBLab dataset was collected in our building at the University of Bonn using a Kuka YouBot platform equipped with several sensors, including a Hokuyo UTM-30LX LiDAR sensor and an up-facing camera.
The up-facing camera is used for determining close to ground truth poses of the robot through localizing densely placed AprilTags on the ceiling that have been measured with a high-precision terrestrial laser scanner~\cite{zimmerman2022iros}.
Additionally, the ground truth poses are optimized by aligning the scans with a highly precise dense point cloud map generated by a Faro terrestrial laser scanner and human-supervised scan matching.
For training our NOF model, we collect a long sequence including $ 31,608 $ frames as the training set.
It consists of several indoor scenes: office, corridor and kitchen, and covers the whole scene.
We additionally collect five shorter sequences for evaluating global localization. 
Each sequence traverses a sub-region of the scene, and the average length of the testing sequences is $ 1419 $ frames.

\textbf{Baselines.} We compare our method with three existing 2D LiDAR-based global localization algorithms:
First, AMCL~\cite{fox2001neurips} as shipped with ROS\footnote{http://wiki.ros.org/amcl}, a widely used highly efficient MCL algorithm; 
Second, the MCL approach by Zimmerman~\etalcite{zimmerman2022iros}, which we call NMCL;
Third, the approach by Sapienza Robust Robotics Group (SRRG), called SRRG-Localizer~\cite{grisetti2018github} shortly SRRG-Loc, which is a sophisticated MCL implemented by the team led by Giorgio Grisetti. 
We additionally re-implemented the HMCL approach using an observation model~\cite{vallicrosa2018h} with a continuous Bayesian Hilbert map~\cite{senanayake2017corl}.
Note that NMCL uses no text spotting to support the localization in our experiments. 
We kept the default parameters for all baseline methods and uses the same number of particles for all approaches for a fair comparison.

\subsection{Global Localization Performance}
\label{subsec:eval_localization}

The first experiment is designed to support the claim that exploits our devised implicit representation-based observation model, our IR-MCL achieves state-of-the-art accuracy in the global localization of a robot.

Our MCL system includes two stages: initialization and pose tracking.
At the initialization stage, we uniformly sample $M = 100,000$ particles in the whole space at the beginning to achieve global localization without any priors.
At the pose tracking stage, we reduce the number of particles to $M = 5,000 $.
HMCL uses the same number of particles as our method. For AMCL, the range of particle numbers is decreased from $ 100,000 $ to $ 5,000 $. We fixed particles number to $ 100,000 $ for NMCL and SRRG-Loc, because they do not provide a mechanism for adjusting the number of particles.
For baseline methods, we build an occupancy grid map or Bayesian Hilbert map by using the data from the training set as the scene representation.
The size of the grid map is $ 50\text{\,m} \times 50\text{\,m} $ with $ 5 $\,cm grid resolution.
For all approaches, we use the same motion model and the odometry reading as control commands.
Thus, the main difference between the different approaches lies in the observation model of the different MCL implementations.

Regarding the evaluation metrics, we calculate the absolute pose error~(APE) between the estimated robot poses and ground truth poses in five testing sets of the IPBLab dataset.
We show both the location error and yaw angle error in terms of the root mean square error~(RMSE) of the location $(x, y)^{\top}$ and the direction $ \theta $ \wrt~the ground truth.
For a fair comparison, we leave out the first $ 20 $\,s as the initialization stage for the MCL and only calculate the localization error when the method converged, \ie,  the initialization stage is excluded for calculating the APE and RMSE.
The localization is regarded as failed if the method cannot converge in $ 20 $\,s, \ie, the location RMSE or yaw RMSE larger than a threshold. We use $ 50 $\,cm for location RMSE and $ 5^\circ $ for Yaw RMSE as a threshold in the experiments.

Besides, we also report the ratio of frames with location and yaw angle errors less than confident thresholds to evaluating the precision of localization results at different tolerance.
The thresholds are 5\,cm, 10\,cm and 20\,cm for the location error, respectively, and $0.5^{\circ}$, $1^{\circ}$, and $2^{\circ}$ for the yaw angle error.
\tabref{tab:ipblab_loc} shows the quantitative results of localization performance.

The experimental results show that our method outperforms the baseline methods in both location and yaw angle accuracy, and that it improves the location accuracy.
As argued before, the main difference between the different methods lies in the observation models, where our method can directly generate a rendered scan from an implicit representation of the scene and does not rely on the discrete occupancy grid map.
In addition, the memory consumption of our NOF representation is only half of the occupancy grid map.
The Hilbert map has a lower memory footprint, but it sacrifices localization accuracy.   
The results show that our method has a good trade-off between performance and memory cost.

\begin{table}[t]
	\centering
	\scalebox{1.0}{
		\begin{tabular}{ccccc}
			\toprule
			\#Particles 	& 10,000 	& 5,000 	& 500 	& 50 \\
			\midrule
			AMCL 	& 10.50/1.52	& 10.45/1.52	& 10.65/1.74	& 18.67/3.44        \\
			NMCL 	& 13.69/2.37	& 13.52/2.27	& 14.06/2.46	& 16.84/3.30        \\
			SRRG-Loc 	& 8.52/1.48  	& 8.48/1.19 	& 8.44/1.54 	& \textbf{9.71/2.24}        \\
			HMCL 	& 21.42/4.90 	& 21.40/4.89	& 20.90/4.84	& 20.84/4.82        \\
			IR-MCL 	& \textbf{6.96/1.14}  	& \textbf{6.85/1.14} 	& \textbf{6.87/1.19} 	& 12.78/2.39      \\
			\bottomrule
		\end{tabular}
	}
	\caption{Ablation study on number of particles in pose tracking stage. We report average APE in Location RMSE\,(cm)/Yaw RMSE\,(degree)  format.}
	\label{tab:particles_number_ablation}
	\vspace{-0.6cm}
\end{table}

\begin{figure}[t]
	\centering
	\includegraphics[width=1.0\linewidth]{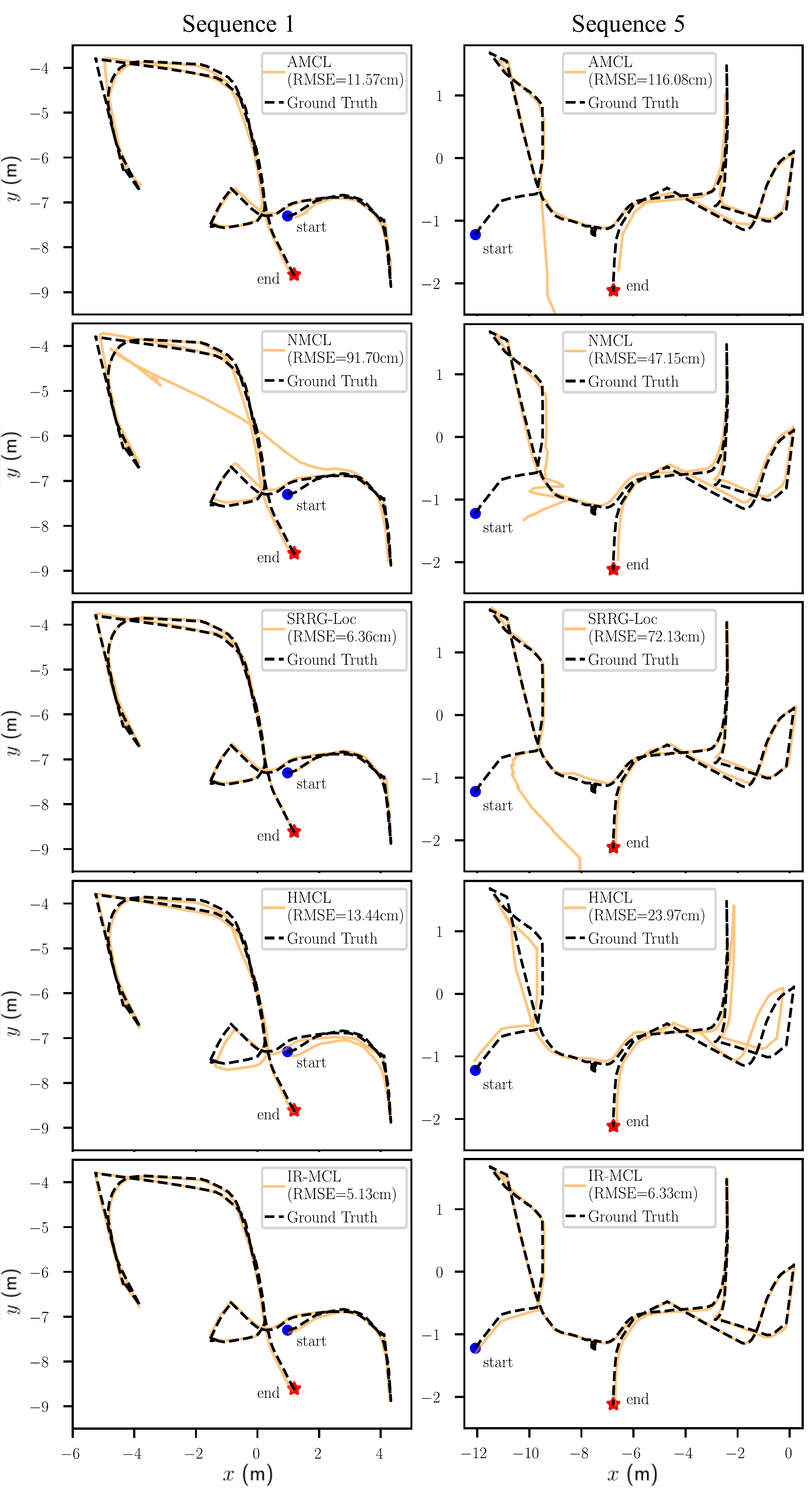}
	\caption{Qualitative global localization results on the sequence 1 and 5 of IPBLab dataset. The dash line depicts ground-truth trajectory, and orange line is the estimated trajectory of the different approaches after convergence. RMSE inside the legend is the location RMSE. Our approach leads to more consistent and more accurate trajectories while achieving faster convergence.}
	\label{fig:loc_results}
	\vspace{-0.6cm}
\end{figure}

\figref{fig:loc_results} shows the qualitative results on sequence 1 and 5 of the IPBLab dataset. We plot the trajectory of each method after the initialization stage, where color indicates the translation error.
We can see that the proposed IR-MCL is much more accurate for global localization as the colors are always in the lower range of the spectrum. 
Furthermore, the figures show that the SRRG-Loc also performs well after convergence, but needs more time in the initialization stage.

\tabref{tab:particles_number_ablation} shows an ablation study on different numbers of particles for pose tracking.
The particles are initialized by adding Gaussian noise to the ground truth pose of the first frame.
The numbers of particles are the same for all methods.
We report the average APE on five sequences of the IPBLab dataset in ``Location RMSE\,(cm)/Yaw RMSE\,($ ^\circ $)" format.
The experimental results show that our method outperforms all baselines in most cases.
SRRG-Loc slightly outperforms our method in the extreme case of only $ 50 $ particles.                  
Moreover, our method stays accurate when particles number change from $ 5,000 $ to $ 500 $. 
Therefore, particles number can be selected according to computing resources in practice.
In this paper, we choose $ 5,000 $ particles to ensure the robustness of our systems after satisfying online operation.

Additionally, the experimental results show that the NMCL and SRRG-Loc work well with fewer particle numbers if provided a good initial guess. However, we still fixed the particle number to $ 100,000 $ in the global localization experiment. It ensures the procedure of these algorithms is not changed for the fairness of the experiment.

\begin{table}[t]
	\centering
	\setlength{\tabcolsep}{3.0pt}
	\renewcommand\arraystretch{1.0}
	\scalebox{1.0}{
		\begin{tabular}{c|c|cccc}
			\toprule
			Dataset                       & Method      & Avg Error\,(m) $\downarrow$ & Acc $\uparrow$ & CD\,(m) $\downarrow$ & F $\uparrow$ \\
			\midrule
			\multirow{2}{*}{Fr079}          & Ray-casting   & 0.33                    & 88.41\%                & 0.17               & 0.97  \\
			& NOF (ours)    & \textbf{0.20}        & \textbf{92.16\%}   & \textbf{0.16}   & \textbf{0.98}  \\
			\midrule
			\multirow{2}{*}{Intel Lab}      & Ray-casting & 0.27                    & 91.62\%                & \textbf{0.19}   & \textbf{0.97}  \\
			& NOF (ours)  & \textbf{0.18}        & \textbf{92.54\%}    & \textbf{0.19}   & \textbf{0.97}  \\
			\midrule
			\multirow{2}{*}{MIT} 			 & Ray-casting  & 0.98         			   & 80.16\%    			 & 0.57   			  & 0.92  \\
			& NOF (ours)   & \textbf{0.45}         & \textbf{81.33\%}    & \textbf{0.37}   & \textbf{0.93} \\
			\bottomrule
		\end{tabular}
	}
	\caption{Quantitative results for the observation model. We compare the rendered scans with the ground truth measurement from the 2D LiDAR. We compare our method (NOF) with the ray-casting methods rendering a scan from an occupancy grid map using Bresenham's algorithm (Ray-casting).}
	\label{tab:observation_model}
	\vspace{-0.6cm}
\end{table}

\subsection{Evaluation of the Observation Model Computed on the Implicit Representation}
\label{subsec:eval_observation_model}

The second experiment is presented to back up the claim that our proposed observation model for a 2D LiDAR sensor based on our proposed implicit representation is more accurate than existing models.
In this experiment, we directly compare rendered scans from our NOF model given poses with the real LiDAR scans.
We take the traditional ray-casting observation model of the beam-end model~\cite{thrun2005probrobbook}, as a baseline, which renders scans using Bresenham's algorithm using an occupancy grid map built by GMapping~\cite{grisetti2007tro}.

We evaluate the performance in Freiburg building 079 dataset, Intel Lab dataset, and MIT CSAIL Lab, which shows that our method is robust in different scenarios.
Because these datasets only contain one sequence, these datasets are not suitable for evaluating global localization.
Here, we split the sequence into subsets for training and evaluation of observation models.

\tabref{tab:observation_model} shows the quantitative evaluation of the observation models.
We use similar error metrics introduced by Rematas~\etalcite{rematas2022cvpr}.
We compare synthesized scans $ \boldsymbol{z} $ with the ground truth scans $ \boldsymbol{\hat{z}} $ and report the average absolute error of the measurements. This metric is the same as the distance function $ \mathcal{D} $ used in our observation model, see \eqref{eq:sensor_model}. Therefore, it directly reflects the errors brought to the localization system.
We also report the accuracy (Acc) as the ratio of the LiDAR beams with a range error smaller than $ 0.5 $\,m compared to the real scan measurements.
Given a LiDAR pose, we can determine the LiDAR beams' origin $ \boldsymbol{o} $ and direction $ \boldsymbol{d} $.
The corresponding 2D point cloud of a scan $ \boldsymbol{z} $ is $ \boldsymbol{p}_{i} = \boldsymbol{o} + z_{i}\boldsymbol{d} $.
We do the same operation for the real scans $ \boldsymbol{\hat{z}} $ to get the ground-truth.
We compute the Chamfer Distance~(CD) and F~-~score~(F) using a threshold of $ 0.5 $\,m between the rendered and the ground-truth point clouds.

As shown in~\tabref{tab:observation_model}, our model generally achieves better performance in all metrics in all datasets.
Our NOF model reduces the average absolute error than the ray-casting method.
The reason is that our method synthesizes more accurate scans compared with the ray-casting method even if there are only training scans with noisy poses available from a SLAM algorithm.
Note that our method significantly improves the accuracy on the MIT dataset by a factor of 2 considering average absolute error.
In this case, it seems that the $ 291 $ frames are not sufficient to build an occupancy grid map to precisely represent the scene. 
But our implicit representation can make reasonable predictions for some places, which are unseen in the training set.
It also supports that our methods have good generalization capabilities for small datasets.
Besides, our method only gets minor improvement in F-score, since it is the harmonic mean of accuracy and completeness.
The results show that our method can reconstruct the complete scene as the occupancy grid map, but is more accurate than~it.

\begin{figure}[t]
	\centering
	\includegraphics[width=1.0\linewidth]{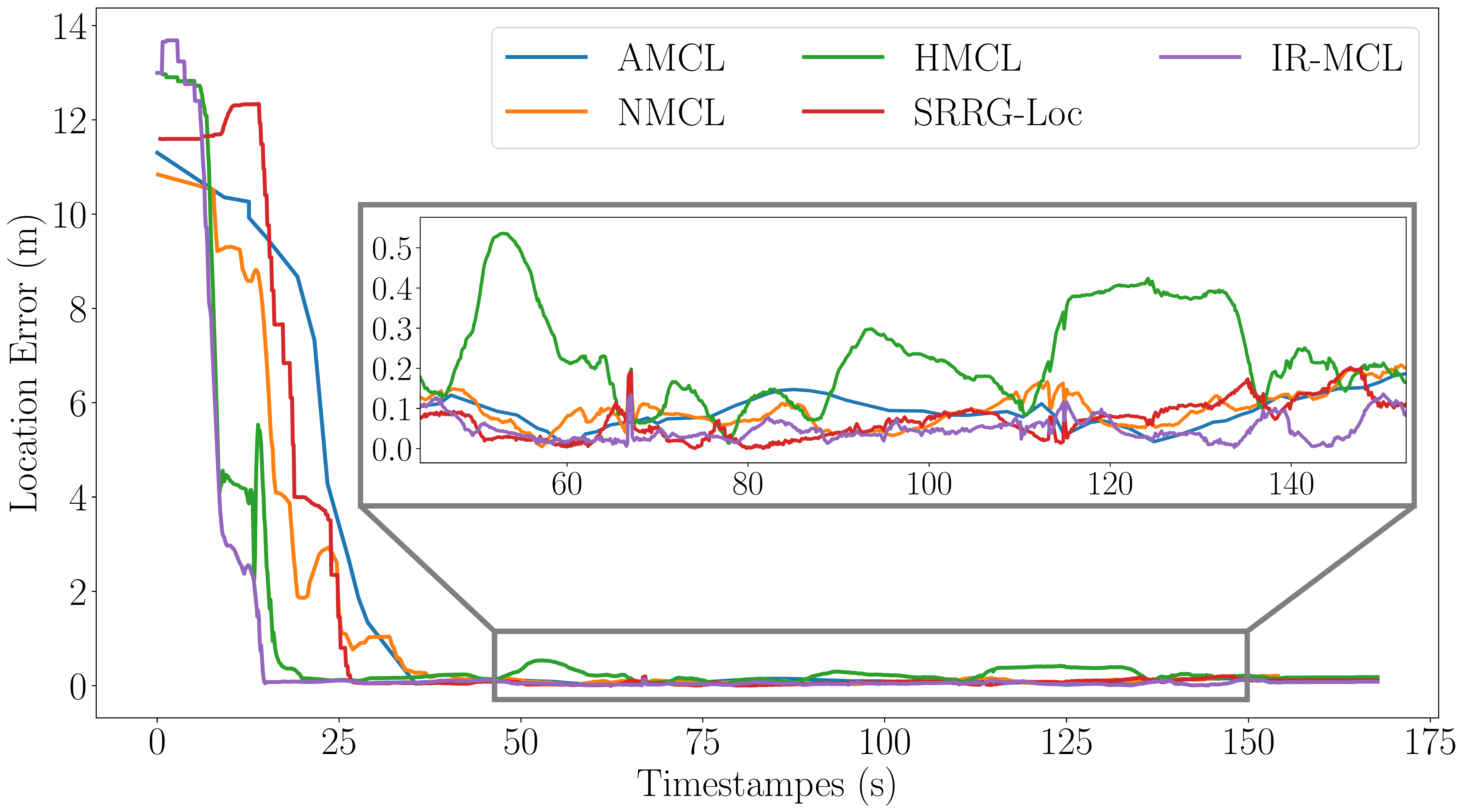}
	\caption{The location error for each frame of sequence 5 of the IPBLab dataset. Our method converges faster than baseline methods, and the localization errors are most of the time lower than the baselines after convergence.}
	\label{fig:converge_curve}
	\vspace{-0.4cm}
\end{figure}

\subsection{Runtime}
\label{subsec:run_time}

The final experiment supports our claim that 
our IR-MCL achieves fast convergence for global localization and can operate online.
We compare our method with all baseline methods on sequence 5 of the IPBLab dataset and we report the location RMSE of every frame of the sequence.
As shown in \figref{fig:converge_curve}, our method converges faster than other baseline methods and achieves higher accuracy after convergence.

We test the runtime of our IR-MCL in initialization stage ($ 100,000 $ particles) and pose tracking stage ($ 5,000 $ particles). 
On a PC with 10 CPU Cores at 3.7\,GHz and 64\,GB memory, and an NVIDIA Quadro RTX 5000 GPU, we achieve an average frame rate of $1.2$\,Hz during initialization stage, and $27$\,Hz after convergence, which supports our claim that computations can be executed fast and in an online fashion.

In summary, our evaluation suggests that we can build an accurate observation model, which provides competitive global localization performance for a mobile robot.
At the same time, our method is fast enough for online processing. 

\section{Conclusion}
\label{sec:conclusion}
In this paper, we presented a novel implicit representation-based online localization approach using a 2D LiDAR.
Our method exploits a neural network-based scene representation to build an accurate observation model.
This allows us to successfully localize a mobile platform in a given environment, and outperform existing gold standard MCL in terms of localization accuracy.
We implemented and evaluated our approach on different datasets and provided comparisons to other existing techniques supporting all claims made in this paper.
The experiments suggest that our approach achieves reliable and accurate global localization while operating online at the sensor frame rate after convergence.
An avenue for future work is to relax the requirement for accurate poses to learn the implicit representation and perform pose estimation at the same time.



\bibliographystyle{plain_abbrv}

\bibliography{glorified,new}

\begin{thebibliography}{10}

\bibitem{adamkiewicz2022ral}
M.~Adamkiewicz, T.~Chen, A.~Caccavale, R.~Gardner, P.~Culbertson, J.~Bohg, and
  M.~Schwager.
\newblock Vision-only robot navigation in a neural radiance world.
\newblock {\em IEEE Robotics and Automation Letters (RA-L)}, 7(2):4606--4613,
  2022.

\bibitem{bennewitz2006euros}
M.~Bennewitz, C.~Stachniss, W.~Burgard, and S.~Behnke.
\newblock {Metric Localization with Scale-Invariant Visual Features using a
  Single Perspective Camera}.
\newblock In H.~Christiensen, editor, {\em European Robotics Symposium 2006},
  volume~22 of {\em STAR Springer Tracts in Advanced Robotics}, pages 143--157.
  Springer Verlag, 2006.

\bibitem{chen2020iros}
X.~Chen, T.~L\"abe, L.~Nardi, J.~Behley, and C.~Stachniss.
\newblock {Learning an Overlap-based Observation Model for 3D LiDAR
  Localization}.
\newblock In {\em Proc.~of the IEEE/RSJ Intl.~Conf.~on Intelligent Robots and
  Systems (IROS)}, 2020.

\bibitem{chen2021icra}
X.~Chen, I.~Vizzo, T.~L\"abe, J.~Behley, and C.~Stachniss.
\newblock {Range Image-based LiDAR Localization for Autonomous Vehicles}.
\newblock In {\em Proc.~of the IEEE Intl.~Conf.~on Robotics \& Automation
  (ICRA)}, 2021.

\bibitem{dellaert1999icra}
F.~Dellaert, D.~Fox, W.~Burgard, and S.~Thrun.
\newblock {Monte Carlo Localization for Mobile Robots}.
\newblock In {\em Proc.~of the IEEE Intl.~Conf.~on Robotics \& Automation
  (ICRA)}, 1999.

\bibitem{deng2022cvpr}
K.~Deng, A.~Liu, J.Y. Zhu, and D.~Ramanan.
\newblock Depth-supervised nerf: Fewer views and faster training for free.
\newblock In {\em Proc.~of the IEEE/CVF Conf.~on Computer Vision and Pattern
  Recognition (CVPR)}, 2022.

\bibitem{fox1999aaai}
D.~Fox, W.~Burgard, F.~Dellaert, and S.~Thrun.
\newblock {Monte Carlo Localization: Efficient Position Estimation for Mobile
  Robots}.
\newblock In {\em Proc.~of the National Conf.~on Artificial Intelligence
  (AAAI)}, 1999.

\bibitem{fox2001neurips}
D.~Fox.
\newblock Kld-sampling: Adaptive particle filters.
\newblock In {\em Proc.~of the Conf.~on Neural Information Processing Systems
  (NeurIPS)}, 2001.

\bibitem{grisetti2018github}
G.~Grisetti.
\newblock srrg-localizer2d (1.6.0).
\newblock \url{https://gitlab.com/srrg-software/srrg_localizer2d}, 2018.

\bibitem{grisetti2007tro}
G.~Grisetti, C.~Stachniss, and W.~Burgard.
\newblock {Improved Techniques for Grid Mapping with Rao-Blackwellized Particle
  Filters}.
\newblock {\em IEEE Trans.~on Robotics (TRO)}, 23(1):34--46, 2007.

\bibitem{he2016cvpr}
K.~He, X.~Zhang, S.~Ren, and J.~Sun.
\newblock {Deep Residual Learning for Image Recognition}.
\newblock In {\em Proc.~of the IEEE Conf.~on Computer Vision and Pattern
  Recognition (CVPR)}, 2016.

\bibitem{hedman2021iccv}
P.~Hedman, P.P. Srinivasan, B.~Mildenhall, J.T. Barron, and P.~Debevec.
\newblock Baking neural radiance fields for real-time view synthesis.
\newblock In {\em Proc.~of the IEEE/CVF Intl.~Conf.~on Computer Vision (ICCV)},
  2021.

\bibitem{ioffe2015icml}
S.~Ioffe and C.~Szegedy.
\newblock {Batch Normalization: Accelerating Deep Network Training by Reducing
  Internal Covariate Shift}.
\newblock In {\em Proc.~of the Intl.~Conf.~on Machine Learning (ICML)}, 2015.

\bibitem{ito2014icra}
S.~Ito, F.~Endres, M.~Kuderer, G.~Tipaldi, C.~Stachniss, and W.~Burgard.
\newblock {W-RGB-D: Floor-Plan-Based Indoor Global Localization Using a Depth
  Camera and WiFi}.
\newblock In {\em Proc.~of the IEEE Intl.~Conf.~on Robotics \& Automation
  (ICRA)}, 2014.

\bibitem{kingma2015iclr}
D.~Kingma and J.~Ba.
\newblock {Adam: {A} Method for Stochastic Optimization}.
\newblock In {\em Proc.~of the Intl.~Conf.~on Learning Representations (ICLR)},
  2015.

\bibitem{li2021iccv}
J.~Li, Z.~Feng, Q.~She, H.~Ding, C.~Wang, and G.H. Lee.
\newblock Mine: Towards continuous depth mpi with nerf for novel view
  synthesis.
\newblock In {\em Proc.~of the IEEE/CVF Intl.~Conf.~on Computer Vision (ICCV)},
  2021.

\bibitem{lu2019cvpr}
W.~Lu, Y.~Zhou, G.~Wan, S.~Hou, and S.~Song.
\newblock {L3-Net: Towards Learning Based LiDAR Localization for Autonomous
  Driving}.
\newblock In {\em Proc.~of the IEEE/CVF Conf.~on Computer Vision and Pattern
  Recognition (CVPR)}, 2019.

\bibitem{maggio2022arxiv}
D.~Maggio, M.~Abate, J.~Shi, C.~Mario, and L.~Carlone.
\newblock Loc-nerf: Monte carlo localization using neural radiance fields.
\newblock {\em arXiv preprint arXiv:2209.09050}, 2022.

\bibitem{mildenhall2020eccv}
B.~Mildenhall, P.P. Srinivasan, M.~Tancik, J.T. Barron, R.~Ramamoorthi, and
  R.~Ng.
\newblock {NeRF: Representing Scenes as Neural Radiance Fields for View
  Synthesis}.
\newblock In {\em Proc.~of the Europ.~Conf.~on Computer Vision (ECCV)}, 2020.

\bibitem{min2022cvpr}
Z.~Min, N.~Khosravan, Z.~Bessinger, M.~Narayana, S.B. Kang, E.~Dunn, and
  I.~Boyadzhiev.
\newblock Laser: Latent space rendering for 2d visual localization.
\newblock In {\em Proc.~of the IEEE/CVF Conf.~on Computer Vision and Pattern
  Recognition (CVPR)}, 2022.

\bibitem{moreau2023wacv}
A.~Moreau, T.~Gilles, N.~Piasco, D.~Tsishkou, B.~Stanciulescu, and
  A.~de~La~Fortelle.
\newblock Imposing: Implicit pose encoding for efficient camera pose
  estimation.
\newblock In {\em Proc.~of the IEEE Winter Conf.~on Applications of Computer
  Vision (WACV)}, 2023.

\bibitem{moreau2021corl}
A.~Moreau, N.~Piasco, D.~Tsishkou, B.~Stanciulescu, and A.~de~La~Fortelle.
\newblock Lens: Localization enhanced by nerf synthesis.
\newblock In {\em Proc.~of the Conf.~on Robot Learning (CoRL)}, 2021.

\bibitem{oechsle2021iccv}
M.~Oechsle, S.~Peng, and A.~Geiger.
\newblock Unisurf: Unifying neural implicit surfaces and radiance fields for
  multi-view reconstruction.
\newblock In {\em Proc.~of the IEEE/CVF Intl.~Conf.~on Computer Vision (ICCV)},
  2021.

\bibitem{ortiz2022rss}
J.~Ortiz, A.~Clegg, J.~Dong, E.~Sucar, D.~Novotny, M.~Zollhoefer, and
  M.~Mukadam.
\newblock isdf: Real-time neural signed distance fields for robot perception.
\newblock In {\em Proc.~of Robotics: Science and Systems (RSS)}, 2022.

\bibitem{o2012ijrr}
S.T. O’Callaghan and F.T. Ramos.
\newblock Gaussian process occupancy maps.
\newblock {\em Intl.~Journal~of Robotics Research (IJRR)}, 31(1):42--62, 2012.

\bibitem{park2019cvpr}
J.J. Park, P.~Florence, J.~Straub, R.~Newcombe, and S.~Lovegrove.
\newblock Deepsdf: Learning continuous signed distance functions for shape
  representation.
\newblock In {\em Proc.~of the IEEE/CVF Conf.~on Computer Vision and Pattern
  Recognition (CVPR)}, 2019.

\bibitem{peng2020eecv}
S.~Peng, M.~Niemeyer, L.~Mescheder, M.~Pollefeys, and A.~Geiger.
\newblock Convolutional occupancy networks.
\newblock In {\em Proc.~of the Europ.~Conf.~on Computer Vision (ECCV)}, 2020.

\bibitem{ramos2016ijrr}
F.~Ramos and L.~Ott.
\newblock Hilbert maps: Scalable continuous occupancy mapping with stochastic
  gradient descent.
\newblock {\em Intl.~Journal~of Robotics Research (IJRR)}, 35(14):1717--1730,
  2016.

\bibitem{rematas2022cvpr}
K.~Rematas, A.~Liu, P.P. Srinivasan, J.T. Barron, A.~Tagliasacchi,
  T.~Funkhouser, and V.~Ferrari.
\newblock Urban radiance fields.
\newblock In {\em Proc.~of the IEEE/CVF Conf.~on Computer Vision and Pattern
  Recognition (CVPR)}, 2022.

\bibitem{senanayake2017corl}
R.~Senanayake and F.~Ramos.
\newblock Bayesian hilbert maps for dynamic continuous occupancy mapping.
\newblock In {\em Proc.~of the Conf.~on Robot Learning (CoRL)}, 2017.

\bibitem{stachniss2003ijcai}
C.~Stachniss and W.~Burgard.
\newblock {Exploring Unknown Environments with Mobile Robots using Coverage
  Maps}.
\newblock In {\em Proc.~of the Intl.~Conf.~on Artificial Intelligence (IJCAI)},
  pages 1127--1132, Acapulco, Mexico, 2003.

\bibitem{stachniss2005aaai}
C.~Stachniss and W.~Burgard.
\newblock {Mobile Robot Mapping and Localization in Non-Static Environments}.
\newblock In {\em Proc.~of the National Conf.~on Artificial Intelligence
  (AAAI)}, 2005.

\bibitem{sucar2021iccv}
E.~Sucar, S.~Liu, J.~Ortiz, and A.J. Davison.
\newblock imap: Implicit mapping and positioning in real-time.
\newblock In {\em Proc.~of the IEEE/CVF Intl.~Conf.~on Computer Vision (ICCV)},
  2021.

\bibitem{thrun2005probrobbook}
S.~Thrun, W.~Burgard, and D.~Fox.
\newblock {\em {Probabilistic Robotics}}.
\newblock MIT Press, 2005.

\bibitem{vallicrosa2018h}
G.~Vallicrosa and P.~Ridao.
\newblock H-slam: Rao-blackwellized particle filter slam using hilbert maps.
\newblock {\em Sensors}, 18(5):1386, 2018.

\bibitem{xu2021neurips}
X.~Xu, X.~Pan, D.~Lin, and B.~Dai.
\newblock Generative occupancy fields for 3d surface-aware image synthesis.
\newblock In {\em Proc.~of the Conf.~on Neural Information Processing Systems
  (NeurIPS)}, 2021.

\bibitem{yan2019ecmr}
F.~Yan, O.~Vysotska, and C.~Stachniss.
\newblock {Global Localization on OpenStreetMap Using 4-bit Semantic
  Descriptors}.
\newblock In {\em Proc.~of the Europ.~Conf.~on Mobile Robotics (ECMR)}, 2019.

\bibitem{yan2021iccv}
Z.~Yan, Y.~Tian, X.~Shi, P.~Guo, P.~Wang, and H.~Zha.
\newblock Continual neural mapping: Learning an implicit scene representation
  from sequential observations.
\newblock In {\em Proc.~of the IEEE/CVF Intl.~Conf.~on Computer Vision (ICCV)},
  2021.

\bibitem{lin2021iros}
L.~Yen-Chen, P.~Florence, J.T. Barron, A.~Rodriguez, P.~Isola, and T.Y. Lin.
\newblock inerf: Inverting neural radiance fields for pose estimation.
\newblock In {\em Proc.~of the IEEE/RSJ Intl.~Conf.~on Intelligent Robots and
  Systems (IROS)}, 2021.

\bibitem{yilmaz2019ras}
A.~Yilmaz and H.~Temeltas.
\newblock Self-adaptive monte carlo method for indoor localization of smart
  agvs using lidar data.
\newblock {\em Journal on Robotics and Autonomous Systems (RAS)}, 122:103285,
  2019.

\bibitem{yuan2018icarcv}
Y.~Yuan, H.~Kuang, and S.~Schwertfeger.
\newblock Fast gaussian process occupancy maps.
\newblock In {\em Proc.~of the Intl. Conf. on Control, Automation, Robotics and
  Vision~(ICARCV)}, 2018.

\bibitem{zhao2020ral}
J.~Zhao, L.~Zhao, S.~Huang, and Y.~Wang.
\newblock 2d laser slam with general features represented by implicit
  functions.
\newblock {\em IEEE Robotics and Automation Letters (RA-L)}, 5(3):4329--4336,
  2020.

\bibitem{zhu2022cvpr}
Z.~Zhu, S.~Peng, V.~Larsson, W.~Xu, H.~Bao, Z.~Cui, M.R. Oswald, and
  M.~Pollefeys.
\newblock Nice-slam: Neural implicit scalable encoding for slam.
\newblock In {\em Proc.~of the IEEE/CVF Conf.~on Computer Vision and Pattern
  Recognition (CVPR)}, 2022.

\bibitem{zimmerman2022iros}
N.~Zimmerman, L.~Wiesmann, T.~Guadagnino, T.~L{\"a}be, J.~Behley, and
  C.~Stachniss.
\newblock Robust onboard localization in changing environments exploiting text
  spotting.
\newblock In {\em Proc.~of the IEEE/RSJ Intl.~Conf.~on Intelligent Robots and
  Systems (IROS)}, 2022.

\end{thebibliography}

\end{document}